\documentclass[a4paper,12pt]{article}

\usepackage[top=2.5cm,bottom=2.5cm,left=2.5cm,right=2.5cm]{geometry}

\usepackage{graphicx}
\usepackage{epstopdf}
\usepackage{subfigure}
\usepackage{amssymb,amsbsy,amsmath,amsfonts,epic,eepic,amsthm}
\usepackage{titling}
\usepackage[numbers]{natbib}
\usepackage{xcolor}
\usepackage{soul}

\begin{document}

\title{Critical Neuromorphic Computing based on\\ 
Explosive Synchronization}

\author{Jaesung~Choi~and~Pilwon~Kim
\thanks{The authors are with Department of Mathematical Sciences, Ulsan National Institute of Science and Technology(UNIST), Ulsan Metropolitan City, 44919,  Republic of Korea. e-mail: pwkim@unist.ac.kr.}}

\maketitle

\begin{abstract}

Synchronous oscillations in neuronal ensembles have been proposed to provide a neural basis for the information processes in the brain. In this work, we present a neuromorphic computing algorithm based on oscillator synchronization in a critical regime. The algorithm uses the high dimensional transient dynamics perturbed by an input and translates it into proper output stream. One of the benefits of adopting coupled phase oscillators as neuromorphic elements is that the synchrony among oscillators can be finely tuned at a critical state. Especially near a critical state, the marginally synchronized oscillators operate with high efficiency and maintain better computing performances. We also show that explosive synchronization which is induced from specific neuronal connectivity produces more improved and stable outputs. This work provides a systematic way to encode computing in a large size coupled oscillators, which may be useful in designing neuromorphic devices.

\end{abstract}

\section{Introduction}

Neuromorphic computing has come to refer to brain-inspired computing architectures to mimic the behaviors of the nerve systems as well as solve challenging machine learning problems\cite{monr, zhao}. At the core level, it comprises a large number of interacting processors\cite{mead} which generates complex dynamics. Such a nonlinear dynamical core, often called a reservoir in neural network community\cite{jaeg,luko}, is perturbed with an external input while a readout layer maps the reservoir dynamics to an output. That is, once the perturbation induces the transient dynamics in a high-dimensional spatio-temporal feature space in the reservoir, the readout layer translates the traces of the system to a target output \cite{solo}. 

In this work, we propose a neuromorphic computing algorithm based on oscillator synchronization. Neurons in the central nervous system behave as nonlinear oscillators, developing rhythmic activity\cite{buzs}. Synchronous oscillations in neuronal ensembles have been proposed to provide a neural basis for the information processes and the coordinated movements\cite{tono, spor, sing, cass}. The Kuramoto model\cite{kura, stro} is a generic model to describe the synchronization phenomena in coupled oscillators and has been used to simulate neuronal synchronization and examine functional implications of the brain connectivity\cite{cumi, kitz, cabr, vill}. 

We are especially interested in how to control the criticality of the oscillator netowrks to maximize the computing performance. It has been observed that the brain operates near a critical state in order to adapt to a great variety of inputs and maximize information capacity\cite{begg1, begg2}. More informative review on the critical dynamics of the brain can be found in \cite{botc}. As such, the neural circuits may achieve extensive computational abilities in a critical regime in which perturbations neither spread nor die out too quickly\cite{del, jaeg}. While there have been some studies on oscillator based computers\cite{fang,pari, torr, yoge2}, the connection to the criticality of the brain have not yet received full attention. One of the benefits of adopting coupled phase oscillators as neuromorphic elements is that one can finely tune the synchrony among oscillators when performing the computing tasks. It can be done by adjusting a coupling strength around a critical value at which a phase transition occurs from incoherency to synchronization. 

Synchronization in a critical regime is also relevant to the neural connectivity. When placed on a network, the Kuramoto model may induce an abrupt and irreversible phase transition in the order parameter as the coupling strength is varied. This explosive synchronization occurs in the specific setting of the natural frequencies of the networked oscillators and the heterogeneous network topologies\cite{gome, leyv}. In this work, we show the effect of topologically induced explosive synchronization on the computing performances.

While the implementation of neuromorphic computing on the device level has been realized in various medium such as oxide-based memristors\cite{maan}, metal insulators\cite{zhou} and spin-torque oscillators\cite{yoge1}, an integrating paradigm for neuromorphic computing is yet far from complete. In this work, we attempt to provide a systematic way to encode computing in a large size coupled oscillators, which may be useful in designing neuromorphic devices.

\section{MODEL}

\subsection{Reservoir of oscillatory networks}

The reservoir of oscillatory networks uses a phase-locked state as the ground state for computations. Once perturbed by inputs, the deviation of the oscillators from the synchronized state is closely observed until they return to the original state. The basic idea underlying the oscillatory reservoir computing is that, if network is large enough, all the information necessary to construct proper computational results can be found in the transient trajectories aroused by inputs.

We consider a network of $N$ neurons, being the dynamics of each of them described by a phase $\theta_i(t) \in [0,2\pi)$:
\begin{equation} \label{eq1}
	\theta_i' = \omega_i + \frac{\lambda_i}{k_i}\sum_{j=1}^{N}A_{ij}\sin(\theta_j-\theta_i), \quad i = 1,\cdots , N,
\end{equation}
where $\omega_i$ is the natural frequency, $\lambda_i>0$ is the coupling strength, and $k_i:=\sum_{j=1}^{N}A_{ij}$ is the degree of the node $i$.  Here $A_{ij}$ is the entry of the adjacency matrix of the network which is equal to 1 if nodes $i$ and $j$ are connected, and zero if they are not. The classical Kuramoto model is defined on the complete graph with an identical coupling strength, that is, $\lambda_i=\lambda$ and $A_{ij}=1$ for all $i$ and $j$. It is commonly observed that a modest coupling strength $\lambda_i>\lambda_{\text{C}}$ in $\eqref{eq1}$ drives the oscillators into a phase-locked state in which they maintain a frozen formation at the same frequency. For given network topology and frequency distribution, one usually is interested in assessment of the critical coupling strength $\lambda_{\text{C}}$  at which a phase transition occurs from incoherency to a phase-locked state.

We can use a measure of synchrony to capture an appropriate coupling strength that leads to the phase-locked states. One measure of synchrony is the Kuramoto order parameter:
\begin{displaymath}
	re^{i\theta} = \frac{1}{N}\sum_{j=1}^{N}e^{i\theta_j}.
\end{displaymath}
The order parameter $r$ achieves its maximum $1$ when the phase of all oscillators are identical in complete phase synchronization. It becomes close to $0$ when the phases are scattered around the circle in dynamical incoherence. The graph(black rectangles) in Figure 1(a)  shows how the magnitude of the order parameter $r$ rises with the coupling strength $\lambda$ in the classical Kuramoto model. The order parameter attains non-zero value for couplings stronger than the critical value $\lambda_{\text{C}}\approx 1.6$, indicating the onset of synchronization.

Increasing the coupling strength in oscillator networks brings the individual frequencies of oscillators one by one to the average frequency of the system until full synchronization is achieved. Recently, in a certain type of oscillator networks\cite{gome, leyv}, discontinuous transitions from incoherent states to phase-locked states have been reported. In those systems, all the effective frequencies persist right up to the synchronization transition and then they suddenly jump to the average frequency simultaneously at the critical point. This phenomenon, called explosive synchronization(ES), was proved to be originated from a positive correlation between the natural frequencies and the coupling strengths of oscillators\cite{zhan}. More specifically, if the coupling strength is proportional to the natural frequency
\begin{equation} \label{eq2}
	\lambda_i=\lambda \lvert \omega_i\rvert,\,\,\, \lambda>0,
\end{equation}
the phase dynamics in $\eqref{eq1}$ induces explosive synchronization. 

The explosive synchronization occurs with hysteresis: besides the forward transition from the incoherent state to the phase-locked states, there is also an abrupt desynchronization with decrease of the coupling strength, which does not overlap with the forward transition. However, we only focus on the forward bifurcation in neuromorphic computing, as we need to keep the system out of the hysteresis loop, avoiding the risk of permanent desynchrony. The plot(red circles) in Figure 1(a) shows that forward discontinuous phase transition occurs at a critical coupling strength $\lambda \approx 2.9$, making striking difference from the continuous phase transition at $\lambda\approx 1.6$.

In this work, we compare the two computing reservoirs based on the Kuramoto model $\eqref{eq1}$ which use the different settings for the coupling strength and the network topology:\\

\indent 1) {\bf Regular synchronization model(RS)} \\
\indent \,\,\,\,\,\,\,\,coupling strength: $\lambda_i=\lambda>0$, \\
\indent \,\,\,\,\,\,\,\,network topology: $A_{ij} = 1$ for all $i$ and $j$.\\
\indent 2) {\bf Explosive synchronization model(ES)} \\
\indent \,\,\,\,\,\,\,\,coupling strength: $\lambda_i=\lambda \lvert \omega_i\rvert\,\,\, \lambda>0,$ \\  
\indent \,\,\,\,\,\,\,\,network topology: a Erd\H{o}s-R\'{e}nyi graph with $1 \leq \langle k_i\rangle <  N$ \\

RS is nothing but the classical Kuramoto model. ES adopts a Erd\H{o}s-R\'{e}nyi graph, which is chosen uniformly at random from the collection of all graphs which have $N$ nodes with a specific mean degree $\langle k_i\rangle$. From here on, we will use networks that consist of $N=500$ oscillators for both models. The natural frequencies $\omega_i$ of oscillators are assumed to follow the normal distribution $N(0,1)$.

\subsection{Choice of coupling strength}

In Kuramoto-based models, a higher coupling strength makes the oscillators synchronized in a tighter phase-locked state. If the synchronization is overly persistent, the transient dynamics induced from the inputs vanish so quickly that it cannot properly handle lengthy computations. On the other hand, under a weak coupling strength, the system may fail to erase the past information which is no more necessary and interferes the current computation as noise. Moreover, the oscillators may not be able to recover the ground state even after the computation is carried out.

For balanced computations, it is reasonable to set the coupling strength at which the phase of the oscillators are marginally locked and can react rapidly to external stimuli. The Kuramoto order parameter $r$ can be used as an indicator for a critical value of the coupling strength $\lambda$. For ES illustrated in Figure 1(a), one may fix the value of $\lambda$ at near $2.9$ where $r$ drops quickly in the backward direction. However, for RS, the phase transition in $r$ gradually arises with $\lambda$ and therefore does not provide a sharp criterion. Moreover, it is not the edge of chaos but the edge of order where the efficient computations occur, in that the system should maintain synchronization as a ground state. Since the conventional Kuramoto order parameter $r$ only yields the continuous phase transition, it is not a good indicator for computational capacity.

In order to overcome this drawback of $r$, we introduce the variance order parameter $r_{\text{var}}$ as
\begin{displaymath}
	r_{\text{var}} = \frac{1}{N}\sum_{j=1}^{N}\exp(-c\, \text{var}_j), \quad c>0
\end{displaymath} 

where $\text{var}_j$ is the temporal variance of the frequency $\theta_j'(t)$. This is a measure for desynchrony that sensitively shows a degree of deviation of oscillators from a steady frequency. For reliable computations, the temporal variance of the frequency should be kept low in the ground state. Note that, for each oscillator, the temporal variance of the frequency becomes 0 if oscillators are in a phase-locked state, keeping their common frequency steady. Figure 1(b) plots $r_{\text{var}}$ for the same formations of the oscillators dealt in 1(a). While $r_{\text{var}}$ almost coincides with $r$ for ES at the critical strength($\lambda \approx 3$), it clearly reveals a discontinuous phase transition for RS which is not observed in $r$. Although $r_{\text{var}}$ does not explicitly show difference between explosive and nonexplosive synchronizations, it provides clear information on a level of the coupling strength for which oscillators are arranged for reliable computations. From here on, we will use the variance order parameter $r_{\text{var}}$ to investigate the relation between the states of the system and its computing performances.

\begin{figure*}[!t]
\centering
\includegraphics[width=0.8\textwidth]{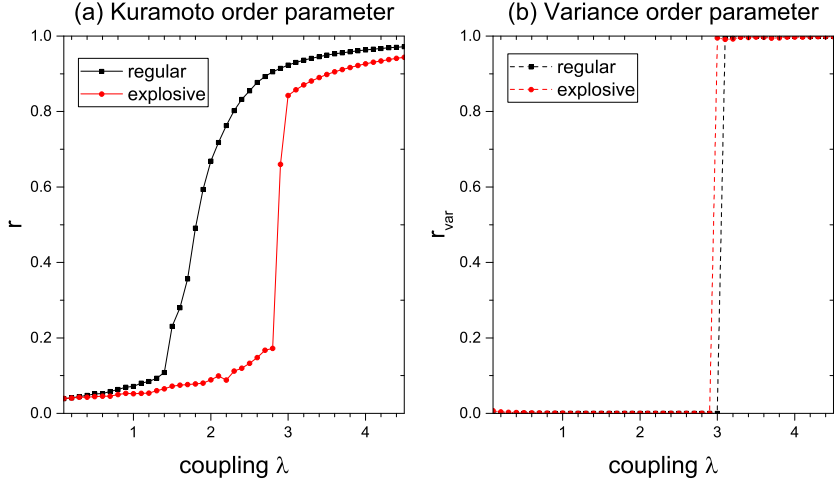}
\caption{Kuramoto and variance order parameters according to the coupling strength: RS and ES use the complete network and a Erd\H{o}s-R\'{e}nyi graph with the mean degree $\langle k_i\rangle=6$, respectively. In ES, only forward transitions are plotted. The parameter $c=10^{-7}$ is used for the variance order parameter.}
\end{figure*}

\subsection{Readout and training}
The oscillator networks are applied to supervised tasks to learn a model that produces a target output $y(t) = (y^1(t), \cdots , y^q(t))\in \mathbb{R}^q$ from an input signal $x(t) = (x^1(t), \cdots , x^p(t))\in\mathbb{R}^p$. In practice the dataset can be either discrete or continuous in time, and also can be multi dimensional signals, but this does not change the principles. The standard training starts with running the network until it reaches a phase locked state. Once oscillators are synchronized, we feed the network the input stream $x(t)$, by forcefully identifying $p$ oscillators in the network with $x(t)$. All evolutionary activities of the oscillators are collected through the frequency values $\theta_i'(t)$ and mapped to the desired output by a trainable readout function $f_{\text{out}}=(f^1_{\text{out}},\cdots,f^q_{\text{out}})\in\mathbb{R}^q$.

In the readout process, to exploit the rich dynamics of the oscillatory networks, it is better to use not only the current states of the oscillators but also their past information. For example, one can constitute a readout function $f^l_{\text{out}},\,l=1,\cdots,q$ as
\begin{displaymath}
	f^l_{\text{out}}(t) = \sum_{i=1}^{N} w_i^l \int_{0}^{t}K(t-\tau)\theta_i'(\tau)d\tau.
\end{displaymath}            
Here $K(t)$ and $w_i^l$ are respectively a kernel function and weights both of which are determined from the training process. However, in spite of the continuous nature of oscillatory networks, we will constitute the readout and training process in this work based on discrete sampling of the signal for simple and clear illustrations. We suggest a readout function that takes past $s$ sampled states of the system at discrete times $t-\Delta t,t-2\Delta t,\cdots,t-s\Delta t$ and maps them to the desired output at time $t$. To be more specific, the readout function $f_{\text{out}}=(f^1_{\text{out}},\dots,f^q_{\text{out}})\in\mathbb{R}^q$ of $(s,\Delta t)$-type is defined as
\begin{equation} \label{eq3}
	f^l_{\text{out}}(t) = \sum_{i=1}^{N}\sum_{j=1}^{s}w^l_{i,j}\theta_i'(t-j\Delta t), \quad l = 1,\cdots,q,
\end{equation}
where $w^l_{i,j}$ are weights to be found from the training process for each computational task. 

The weights $w^l_{i,j}$ are determined so that $f_{\text{out}}(t)$ matches $y(t)$ as close as possible, minimizing an error measure. For example, if the available output data is a time series of total length $M$, $y(t_1),y(t_2),\cdots,y(t_M)$, a typical mean-square error is
\begin{equation} \label{eq4}
	\frac{1}{M}\sum_{i=1}^{M} \lVert y(t_i)-f_{\text{out}}(t_i)\rVert^2 .
\end{equation}

\section{Numerical tests}

In the following numerical examples to test the learning ability of the oscillator networks, we use the $(10,0.1)$-type readout function: The output $f_{\text{out}}(t)$ at $t$ is obtained from $10$ previous sampled values of the oscillator frequencies $\theta'_i(t-0.1),\cdots,\theta'_i(t-1)$ through the equation $\eqref{eq3}$. 

We set up two tasks, filtering and forecasting, both of which require the presence of long-term memory for proper execution.  Task 1 is to learn the scalar output
\begin{displaymath}
	y(t) = \frac{1}{m}\sum_{k=1}^{m}\left( a x(t-k)+b x(t-k)^2+c x(t-k)^3 \right)
\end{displaymath}
which is determined from the past $m$ values of an input stream $x(t)$. Here $a,b$ and $c$ are some nonzero parameters. We use the input $x(t)$ generated from the Lorenz system which provides standard benchmark task for chaotic series handling\cite{casd}. Note that, if $m=1$, the task is simply to implement a polynomial function of the current value of the input. The task becomes more challenging as $m$ increases, requiring long-term memory to evaluate averaged values.

Task 2 is the time series prediction. Based on a previous input stream of $x(t)$, the network is required to predict $m$ steps ahead, that is, the next $m$ values, $x(t+1),\cdots,x(t+m)$. This implies that the desired output vector $y(t) = (y^1(t),\cdots,y^m(t))\in\mathbb{R}^m$ at time $t$ satisfies $y^l(t) = x(t+k), \, k=0,\cdots,m-1$. We take the input $x(t)$ from Mackey-Glass equation which is a chaotic time-delayed differential equation.

In each task, the continuous input signal $x(t)$ and the target signal $y(t)$ are generated for $t\in [0,5000]$. The training process is applied to match $f_{\text{out}}(t)$ to y(t) over the first 4,000 discrete time steps, $t=1,2,\cdots,4000$. That is, the readout weights $w^l_{ij}$ in $\eqref{eq3}$ are determined to minimize the averaged error $\eqref{eq4}$ with respect to $M=4000$. Then we measure the performance using the remaining part of the signal for $t\in(4000,5000]$: the averaged error $\eqref{eq4}$ between $f_{\text{out}}(t)$ to $y(t)$ is evaluated over 1,000 discrete sampled time steps. 

\subsection{Computing performance at the critical point}
We first illustrate the computational performance of RS on complete networks. Figures 2 (a) and (b) respectively reports the averaged errors in task 1 and 2. We measure the errors brought by the change of the coupling strength $\lambda$. It is observed that, in the both tasks, the errors is minimized at the common point which coincides with the critical point in $r_{var}$ in Figure 2(c). Note that $\lambda=3$ indicates where the desynchronization begins. One can confirm that the computational capability of RS attains its maximum at the edge of the synchronization, regardless of the task length($m=5,10$ and $15$) and types(filtering and predictions).

\begin{figure}[!t] 
	\centering
	\includegraphics[width=0.5\textwidth]{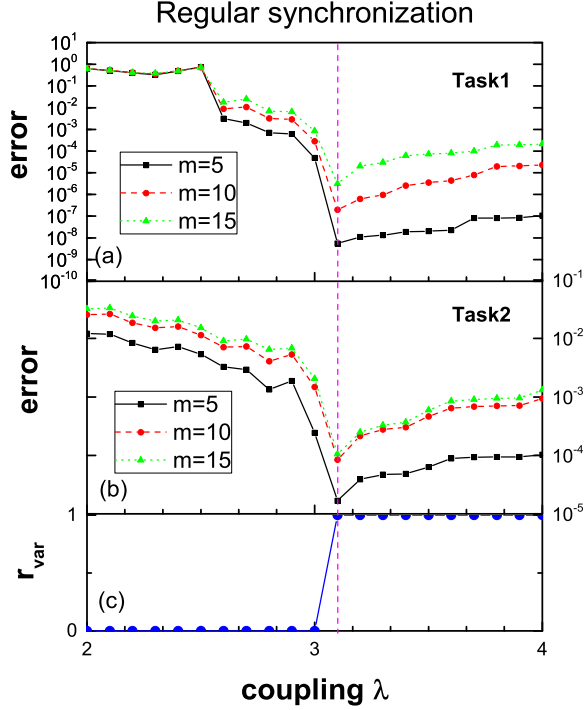}
	\caption{Test error according to the coupling strength in RS. The sudden changes in error in (a) and (b) coincides with the criticality in the variance order parameter in (c).}
\end{figure}

\subsection{Computing with explosive synchronizations}

Having observed the optimized computing ability of RS at criticality, we now turn to the case where the critical transition occurs with the explosive synchronization. We apply ES on a Erd\H{o}s-R\'{e}nyi graph with the mean degree $\langle k_i\rangle=6$. Figure 3 shows that the test errors in the both tasks drop at the common coupling strength, likewise in the case of RS. However, one can see that the accuracy has improved significantly by $10$ to $1000$ times, compared to those of RS. 
Another observation is that the error level maintains even for stronger coupling forces beyond the critical point, while it slowly increases in RS in Figure 2. 

\begin{figure}[!t] 
	\centering
	\includegraphics[width=0.5\textwidth]{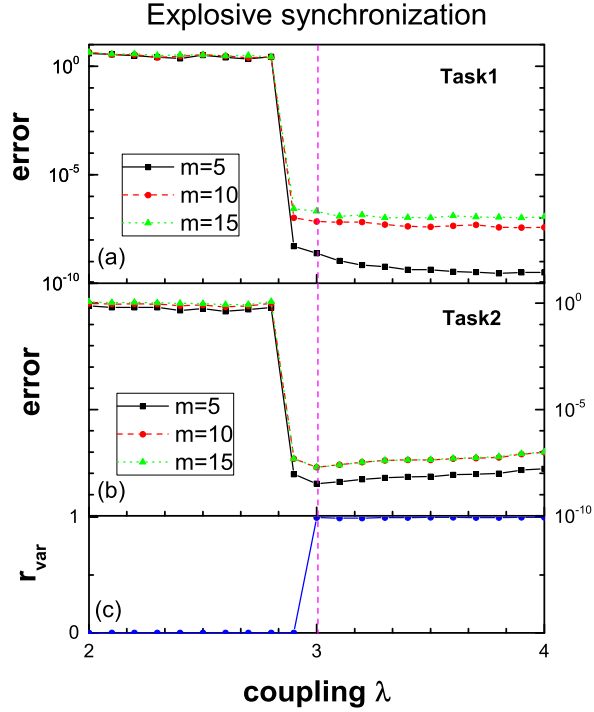}
	\caption{Test error according to the coupling strength in ES. The sudden changes in error in (a) and (b) coincides with the criticality in the variance order parameter in (c).}
\end{figure}

It is assumed that the computing ability to deal with various input signals in different tasks is closely related to the spectral properties of the system reacting to perturbations. Since individual oscillators in ES hold their own effective frequencies until they turn to have the same effective frequency at the onset of synchronization\cite{gome}, frequencies of various modes undergo the same criticality. This implies that the system is well prepared for different tasks which involve a wide range of wavelengths. Figure 4 compares the errors of RS and ES according to the task length. While the error of RS sharply increases with the task length, the performance of ES maintains a descent accuracy level in both tasks.

\begin{figure*}[!t] 
	\centering
	\includegraphics[width=0.8\textwidth]{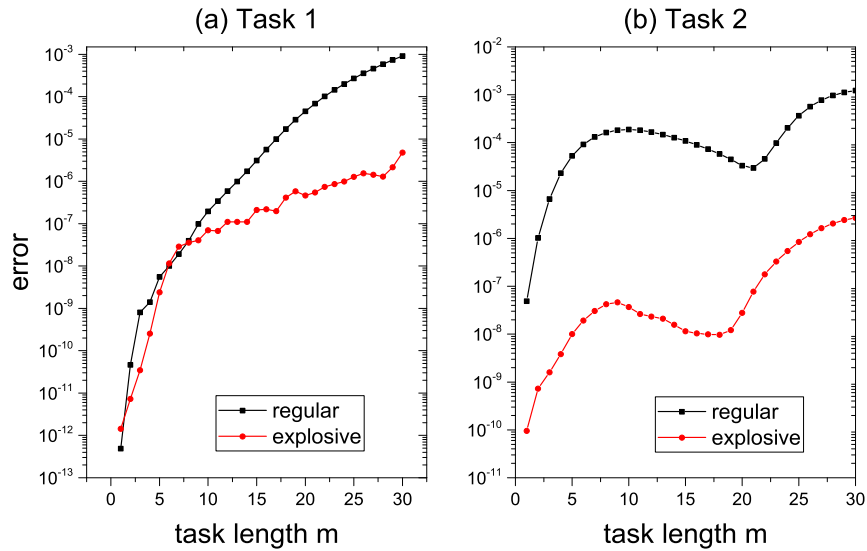}
	\caption{Test error according to the task length for RS and ES}
\end{figure*}

In order to investigate the spectral sensitivity of the systems, we additionally test with an input signal of various frequency modes,
\begin{equation} \label{eq5}
	x(t) = \frac{1}{m}\sum_{i=1}^{m}(a_i\sin(b_it+c_i)+d_i),
\end{equation}
where $a_i,b_i,c_i,d_i \in \mathbb{R}, i = 1,\cdots,m.$

In Figure 5, the errors of ES grow relatively slower than those of RS as the number of frequency modes $m$ increases.

\begin{figure*}[!t] 
	\centering
	\includegraphics[width=0.8\textwidth]{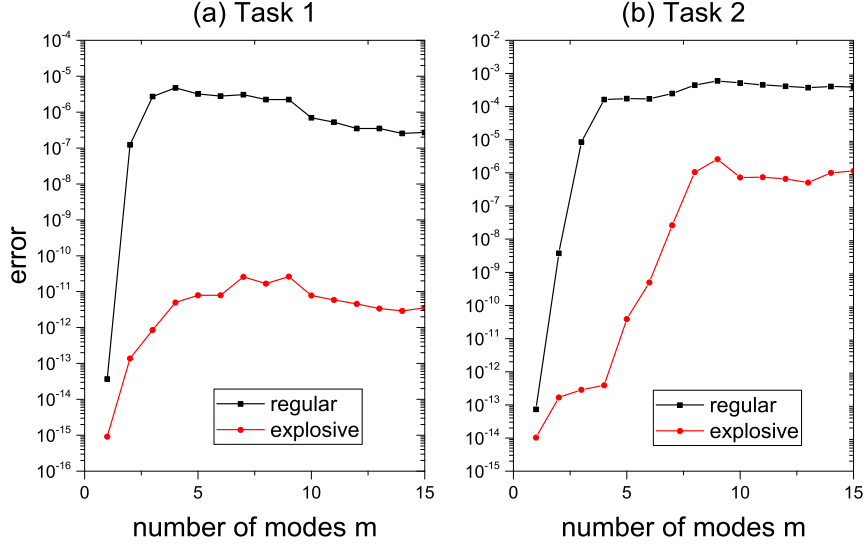}
	\caption{Test error according to the number of modes of frequency in the input stream $\eqref{eq5}$ for RS and ES}
\end{figure*}

\subsection{Clear reset in sparse networks}

The explosive synchronization can occur on networks with a variety of topological structures, as long as the frequency-coupling relation $\eqref{eq2}$ holds\cite{zhan}. In this section, we investigate ES in the random networks with various levels of connectivity. Figure 6 illustrates the change of the test errors with respect to the mean degree of Erd\H{o}s-R\'{e}nyi graphs. One can see that the performances in two tasks are minimized when the mean degree is at about 6 to 24. If the networks are too sparse, say $\langle k_i\rangle\leq3$, they are likely to form separate sub-networks, making close cooperation of oscillators impossible. On the contrary, it is noted in Figure 6 that the densely connected oscillators do not work well either. The errors slowly increase with the mean degree as viewed.

\begin{figure*}[!t] 
	\centering
	\includegraphics[width=0.8\textwidth]{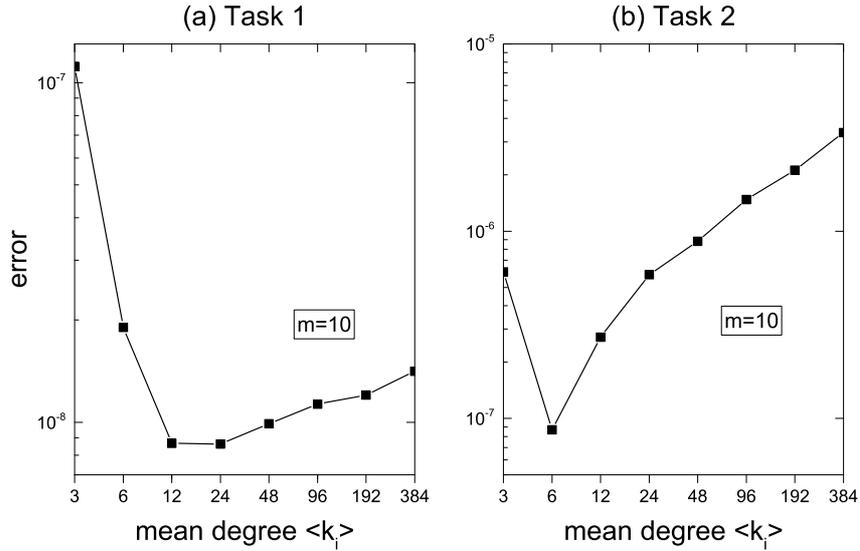}
	\caption{Test error according to the mean degree in ES}
\end{figure*}

These phenomena can be understood in terms of the reset mechanism in computations. Once outputs are generated from transient dynamics induced by inputs, the system should bring its elements to normal condition or the initial state. This is necessary for the system to prepare for next inputs and produce reliable results. In oscillatory networks, a phase-locked state plays a role of this ground state. If the oscillators are densely connected, there are likely excessive ensembles of such ground state. A large number of possible initial states can weaken the capability to reproduce consistent results.

\subsection{Phase transition in training error}

The order parameter needs to be evaluated first for tuning the systems at the critical regime. However, evaluation of the order parameter could be expensive, even impossible, if access to all oscillators are not feasible. A practical alternative is to measure the training error from the outputs instead of evaluating the order parameters. Figure 7 shows that the training error of RS sharply drops at the same critical point as in Figure 2. That is, one can detect a critical coupling strength from a sudden change in the training error. Note that the training error is kept as high as the test error until the coupling strength reaches a certain level, which implies that RS hardly learns from the training set in a weak coupling regime.

\begin{figure*}[!t] 
	\centering
	\includegraphics[width=0.8\textwidth]{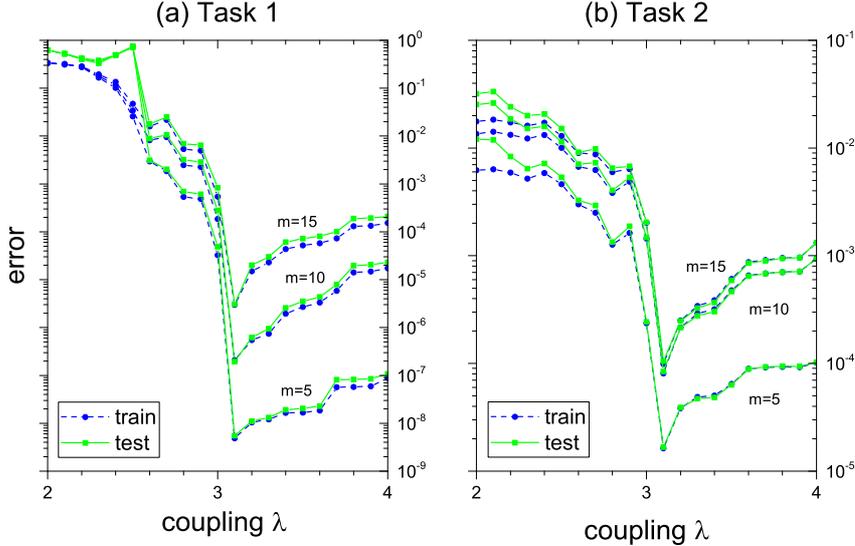}
	\caption{Training and test error according to the coupling strength in RS. The sudden changes at $\lambda \approx 3$ coincide with the criticality in the variance order parameter.}
\end{figure*}

Interestingly, in Figure 8, the more dramatic and exactly opposite situation occurs with ES: the training error keeps low and makes a sharp rise at the critical point. A weak coupling strength in ES leads to overfitting to training data set and lack of generalization to handle new inputs. In contrast to RS, a sudden rise in the training error in ES indicates that the coupling strength reaches a critical level for synchronization.

\begin{figure*}[!t] 
	\centering
	\includegraphics[width=0.8\textwidth]{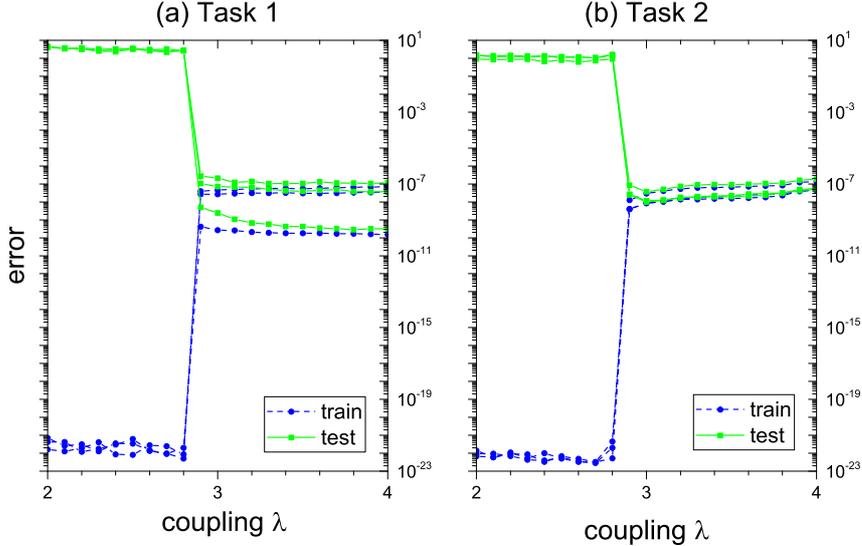}
	\caption{Training and test error according to the coupling strength in ES. The sudden changes at $\lambda \approx 3$ coincide with the criticality in the variance order parameter.}
\end{figure*}

\section{Discussion}

With a low coupling strength, oscillators are not able to form a consistent initial state from where valid computation can start. On the contrary, an excessive coupling strength suppresses dynamics perturbed by external stimuli too quickly, preventing it from working for efficient computation. Simulations showed that networks of phase oscillators maximize their dynamic range of information processing when configured on the edge of the synchronization. They can provide a general framework for neuromorphic computing in that their synchronoy can be easily controlled by the coupling strength. Especially in the explosive synchronization, in a critical parameter regime where every mode of frequencies undergoes simultaneous synchronization, the computing performance is greatly improved compared to that of the regular synchronization. 

Since our models consist of a large number of coupling oscillators, a less number of connections leads to lower computational cost and more economical implementation of machines. Simulations revealed that sparse oscillatory networks generally work better, as long as they are placed at a critical regime for explosive synchronization. This further suggests that the number of the oscillators that are directly referred by the output components can be also effectively reduced. 

To find the critical coupling strength, we used the variance order parameter that clearly indicates the onset of phase-locking by discontinuous jump in either regular or explosive synchronization. However, since evaluation of the variance order parameter needs a long-time access to the entire network, measuring it can be impractical, if not impossible. We showed that tracking the training error can replace the order parameter: one can increase the coupling strength until there appears a sudden change (up/down) in the training error. Since evaluating the training error is a part of every learning process, we can locate the critical coupling strength without additional cost. The opposite directions of the phase transitions observed from regular and explosive synchronization is left for future studies.

\section*{Acknowledgment}
This work was supported by the Ministry of Education of the Republic of Korea and the National Research Foundation of Korea (NRF-2017R1D1A1B04032921). The funder had no role in study design, data collection and analysis, decision to publish, or preparation of the manuscript.

\bibliographystyle{plain}
\bibliography{CNC_references}

\end{document}